\documentclass[letterpaper, 10 pt, conference]{ieeeconf}  

\IEEEoverridecommandlockouts                              

\usepackage{graphicx} 
\usepackage{multirow}
\usepackage{booktabs}
\usepackage{amsmath}
\usepackage{amsfonts}
\usepackage{algorithm}
\usepackage{algpseudocode}
\usepackage{bm}
\usepackage[colorlinks,linkcolor=red]{hyperref}
\usepackage{xcolor}
\title{\LARGE \bf
SG-Bot: Object Rearrangement via\\ Coarse-to-Fine Robotic Imagination on Scene Graphs
}

\author{Guangyao Zhai$^1$, Xiaoni Cai$^1$, Dianye Huang$^1$, Yan Di$^{1,\dagger}$\\ Fabian Manhardt$^{2}$, Federico Tombari$^{1,2}$, Nassir Navab$^{1}$ and Benjamin Busam$^{1}$ \\ [0.5em]
\url{https://sites.google.com/view/sg-bot}
\thanks{}
\thanks{$^{1}$ Technical University of Munich. $^{2}$ Google. $^{\dagger}$ Corresponding author.}
}

\algnotext{EndFor}
\algnotext{EndIf}

\begin{document}

\setcounter{figure}{1}
\makeatletter
\let\@oldmaketitle\@maketitle
\renewcommand{\@maketitle}{\@oldmaketitle
  \begin{center}
    \includegraphics[width=\textwidth]{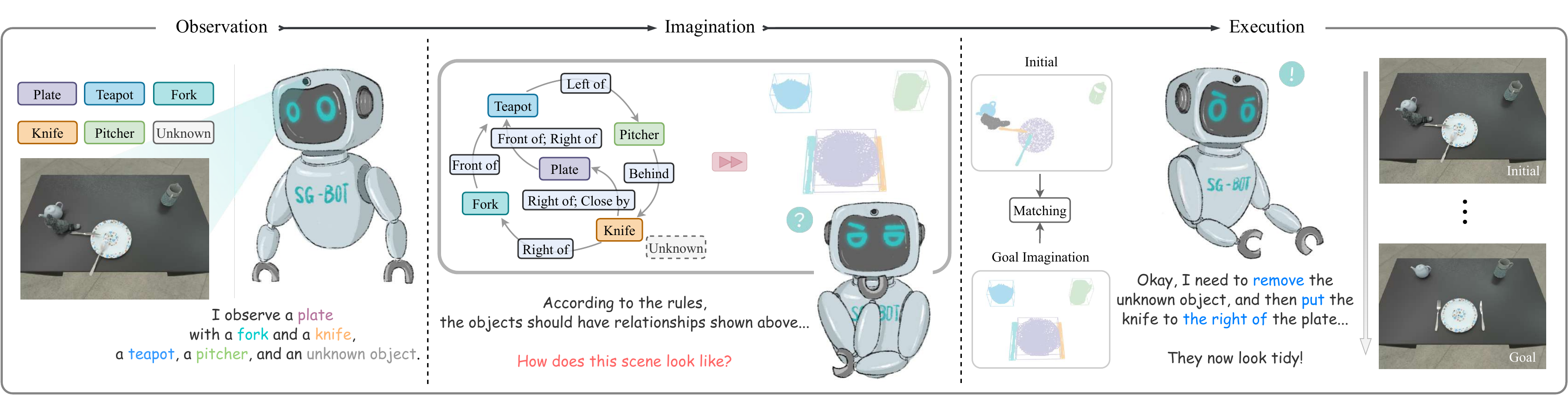}
    \label{fig:cover}
  \end{center}
  \vspace{-15pt}
  \footnotesize{Fig.~\thefigure.~\label{fig:cover}~\textbf{SG-Bot workflow.}
  SG-Bot designs a three-fold procedure for robotic rearrangement.
  The cluttered initial scene is first perceived and processed into individual object nodes in \textit{Observation}. 
  Then, these objects are transitioned into the \textit{Imagination} phase, where scene graph representation is adopted to facilitate the coarse-to-fine goal scene imagination, fusing all available priors and user commands.
  Finally, robotic action policies are generated in \textit{Execution} by matching the initial and goal scenes.
  SG-Bot is lightweight, real-time, and controllable.}
  \label{fig:cover}
 \vspace{-15pt}
  \medskip}
\makeatother

\maketitle
\thispagestyle{empty}
\pagestyle{empty}
\begin{abstract}
Object rearrangement is pivotal in robotic-environment interactions, representing a significant capability in embodied AI.
In this paper, we present SG-Bot, a novel rearrangement framework that utilizes a coarse-to-fine scheme with a scene graph as the scene representation.
Unlike previous methods that rely on either known goal priors or zero-shot large models, SG-Bot exemplifies lightweight, real-time, and user-controllable characteristics, seamlessly blending the consideration of commonsense knowledge with automatic generation capabilities.
SG-Bot employs a three-fold procedure--observation, imagination, and execution--to adeptly address the task.
Initially, objects are discerned and extracted from a cluttered scene during the observation. 
These objects are first coarsely organized and depicted within a scene graph, guided by either commonsense or user-defined criteria.
Then, this scene graph subsequently informs a generative model, which forms a fine-grained goal scene considering the shape information from the initial scene and object semantics.
Finally, for execution, the initial and envisioned goal scenes are matched to formulate robotic action policies.
Experimental results demonstrate that SG-Bot outperforms competitors by a large margin.
\end{abstract}
\section{Introduction}
Object rearrangement is an essential but challenging task in robot-environment interaction, marking a crucial capability in embodied AI~\cite{batra2020rearrangement}. 
This interactive ability attains its zenith of automation by synergizing vision~\cite{gu2021open,carion2020end,kirillov2023segment,di2021so}, textual insights from sources~\cite{brown2020language,chowdhery2022palm,jiang2022vima}, and strategic motion planning~\cite{li2023behavior,ding2023task}. 
Together, these elements culminate in a sophisticated physical embodiment for robots.

Robotic rearrangement refers to the process wherein a robotic agent, starting from an initial configuration within a scene, re-positions objects according to specific rules or instructions.
The purpose is to achieve desired goal states, relying solely on sensory data and onboard perceptions.
Recently proposed vision-based solutions to this task can be categorized into three approaches: \textbf{utilizing known geometric and semantic goal states}, \textbf{sequential object pose estimation}, and \textbf{zero-shot rearrangement with large models}.
Typically, for goal-guided methods~\cite{goyal2022ifor, goodwin2022semantically}, the quality of such priors significantly affects the accuracy of the rearrangement. 
When the goal state is unavailable, such methods become inapplicable for real-world use.
Moreover, for pose estimation based approaches~\cite{liu2022structformer}, while the sequential design aligns well with robotic manipulations, it can be affected by cumulative errors in autoregressive predictions.
The last type of methods~\cite{kapelyukh2023dall,ahn2022can,zeng2022socratic,brohan2023rt,driess2023palm} tap into commonsense knowledge stored in zero-shot models.
They necessitate either intricate post-filter procedures or prompt template designs, which tend to overlook scene-specific contextual cues and result in diverse undesired outcomes. 

Orthogonal to the above methodologies, we explore a novel rearrangement routine embodied as \emph{SG-Bot}, using goal imagination on scene graphs and goal-guided object matching as shown in Fig.\hyperref[fig:cover]{~\thefigure}. SG-Bot stacks three stages for the task, which are \textit{observation}, \textit{imagination}, and \textit{execution}. Specifically, in the first stage, it processes initial scenes to extract objects by semantic instance segmentation. The imagination stage follows a coarse-to-fine solution, where objects are firstly treated as semantic nodes in a constructed goal scene graph. This graph is either directed by commonsense reasoning or user-defined rules, serving as coarse goal states. For a finer generation, the goal scene graph can already be decoded to an actual scene using a scene generative model, Graph-to-3D~\cite{dhamo2021graph}. 
However, inherited from the features of generative models, Graph-to-3D can produce diverse generation results inconsistent with the observation, potentially affecting the precision of subsequent object matching. We control the generation process by enriching the graph with shape priors to make a shape-aware graph, equipping the initial shape knowledge. Next, SG-Bot performs finer goal scene imagination conditioned on this graph, ensuring that the imagined shapes are coherent with the initial observation. Finally, in the execution stage, the imagined objects serve as anchors to guide the object matching by point cloud registration during the scene transformation. At each transformation step, we check occupancy between objects in the current observation and the imagination for safe rearrangement. The uniqueness of SG-Bot manifests in three aspects: First, SG-Bot does not need known goal priors but can self-generate goal scenes exclusively for the initial scenes, compared to the goal-required methods, \textit{e.g.},~\cite{goyal2022ifor,goodwin2022semantically}. Second, SG-Bot decouples the transformation policy using per-object matching to decrease the risk of error accumulation, compared to autoregressive methods, \textit{e.g.},~\cite{liu2022structformer}. Third, the concrete goal states and the closed-loop rearrangement strategy guarantee the rearrangement performance, compared to the loose-coupled zero-shot methods, \textit{e.g.},~\cite{zeng2022socratic}.

Our contributions are summarized as:
\begin{itemize}
\item We present \emph{SG-Bot}, a new paradigm for the object rearrangement. The goal states are coarse-to-fine generated on the rules represented as scene graphs, with which goal-guided matching defines our motion policies.
\item Ambiguous goal scene generation is alleviated by extracting shape priors from the initial observation. This leads to improved rearrangement performance.
\item Experimental results in simulation show that SG-Bot can achieve competitive performance with state-of-the-art methods. Moreover, the rearrangement performance remains consistent in real-world scenarios.
\end{itemize}
\section{Related Work}

\subsection{Scene Graph}
Scene graphs offer a rich symbolic and semantic representation of scenes~\cite{chang2021comprehensive,johnson2015image}. They can reason about objects and their relationships more explicitly than language~\cite{johnson2018image}. This compact relationship description can be obtained through spatial grounding~\cite{wald2020learning,armeni20193d}, predicted from images~\cite{dhamo2020semantic,wu2021scenegraphfusion,xu2017scene}, or even a GUI~\cite{sggui}. Scene graphs have applications in numerous computer vision areas such as 2D image generation~\cite{yang2022diffusion,johnson2018image}, manipulation~\cite{dhamo2020semantic}, caption generation~\cite{krishna2017visual}, camera localization~\cite{rosinol2021kimera}, and 3D scene synthesis~\cite{luo2020end,dhamo2021graph,zhai2023commonscenes}. Recent robotics manipulation research also leverages scene graphs in planning~\cite{tang2023selective,zhu2021hierarchical,rana2023sayplan}. In the context of this work, scene graphs serve to generate scenes, acting as anchors that guide the rearrangement.

\subsection{Object Rearrangement}
The task necessitates that an embodied agent transition from initial states to goal states, adhering to specific rules based on perception and planning~\cite{batra2020rearrangement}, as indicated by earlier works~\cite{cosgun2011push,king2016rearrangement,king2017unobservable,lee2019efficient,coumans2021}.  By leveraging the development of visual perception~\cite{wang2019densefusion,peng2019pvnet,manhardt2019explaining,zhang2022ssp,di2022gpv}, robotic grasping~\cite{zhai20222,sundermeyer2021contact,zhai2023monograspnet}, motion planning~\cite{wang2021uniform,cheong2020relocate,gao2022fast}, and research platforms~\cite{savva2019habitat,szot2021habitat,kolve2017ai2,james2020rlbench,xiang2020sapien,shen2021igibson}, a number of related methods have emerged. Solutions for this task fall into two categories. First, the goal states are given to the embodied agent, subsequently solving the problem by object matching, for example, using optical flow~\cite{goyal2022ifor} or feature cosine similarity~\cite{goodwin2022semantically}. However, deriving such configurations can be challenging in real-world scenarios. Secondly, the goal states can be generated conditioned on the initial states. These goal states can be implicitly represented, such as by gradient fields~\cite{wu2022targf}, scene distributions~\cite{wei2023lego}, or sequential reasoning on the observation~\cite{liu2022structformer}. Alternatively, goals can be explicit in various formats, such as images~\cite{kapelyukh2023dall} on prompts, bounding boxes~\cite{gkanatsios2023energy} or poses~\cite{kapelyukh2023dream2real} on descriptions, and direct language instructions~\cite{ahn2022can,singh2023progprompt,liang2023code}, leveraging recent off-the-shelf large language models~\cite{ramesh2022hierarchical,brown2020language,chen2021evaluating}. More powerful models even treat the initial-goal transformation as an end-to-end problem~\cite{brohan2022rt,brohan2023rt}, building on the large resource consumption. In this work, we generate the goal in a two-stage fashion, where coarse relationships are symbolized as a scene graph and finer concrete goals as the imagined scene given by the scene graph.
\section{Preliminary}

\begin{figure*}[t]
    \centering
    \includegraphics[width=1.0\textwidth, angle=0]{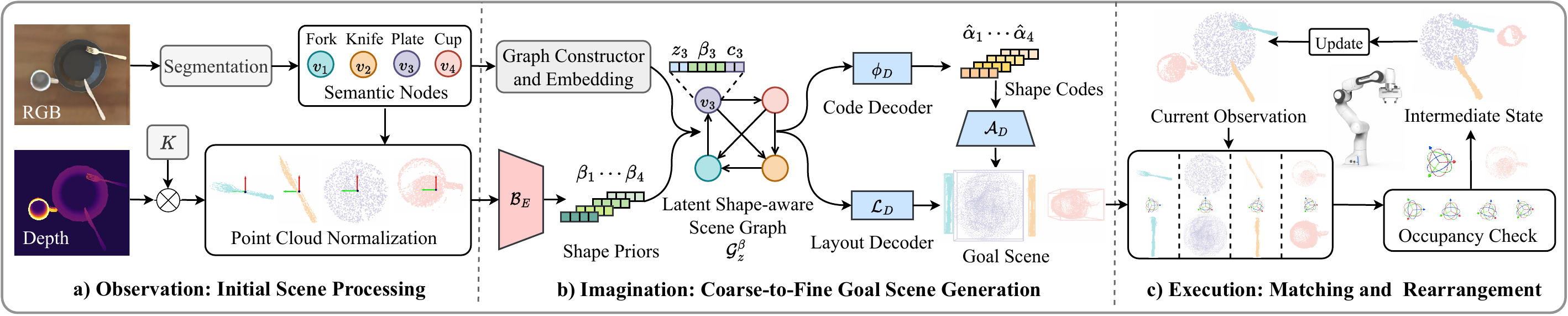}
    \vspace{-15pt}
    \caption{\textbf{SG-Bot pipeline.}
    \textbf{a)} 
    SG-Bot segments the input RGB image via MaskRCNN~\cite{he2017mask} to obtain individual object nodes $v_i$.
    Then, the corresponding point cloud of $v_i$ is obtained via back-projecting the depth map with camera intrinsics $K$.
    \textbf{b)} Coarse: the graph constructor connects each pair of nodes according to commonsense or user-defined rules, yielding scene graph $\mathcal{G}$.
    Fine: $\mathcal{G}$ is embedded and enhanced to $\mathcal{G}_z^\beta$ by combining estimated shape priors $\beta^*$ extracted from the normalized point clouds using the trained encoder $\mathcal{B}_E$ and latent code $z$ sampled from the learned layout-shape distribution.
    $\mathcal{G}_z^\beta$ then informs $\mathit{\Phi}_D$ and $\mathcal{L}_D$ of Graph-to-3D~\cite{dhamo2021graph} to generate shape codes $\alpha^*$ and the scene layout respectively. $\alpha^*$ are decoded as shapes via $\mathcal{A}_D$, which are then populated in the layouts to form the goal scene.
    \textbf{c)} SG-Bot matches the initial and envisioned goal using point cloud registration and performs an occupancy check to determine the final movement in each step, as illustrated in~\ref{matching}. The robot iteratively executes the action, transforming scenes into intermediate states and updating the observation until it reaches the goal state.
    }
    \label{fig:workflow}
    \vspace{-0.4cm}
\end{figure*}%

\textbf{Scene Graph.}
\label{scenegraph}
The scene graph we use is semantic scene graph~\cite{chang2021comprehensive}, denoted as $\mathcal{G} = \left\{\mathcal{V}, \mathcal{E}\right\}$, which serves as a structured representation of a visual scene.
In such representation, $\mathcal{V} = \{v_{i}~|~i = {1, \ldots, N \} }$ refers to the set of object nodes, while $\mathcal{E} = \{e_{i \to j}~|~i,j = {1, \ldots, N}, i \neq j \}$ represents the set of directed edges connecting each pair of nodes $v_{i} \rightarrow v_{j}$. 
As structured in the left of Fig.~\ref{fig:train}.b, each node $v_{i}$ can encompass various extensible attributes, \textit{e.g.}, object category information $o_{i} \in O$, with $O$ containing all categories. 
As same as the node representation, each edge $e_{i\to j}$ is associated with a class label $\gamma_{i\to j} \in \Gamma$.
In this paper, $\Gamma$ contains all pre-defined edge types, \textit{i.e.}, $\{$\texttt{left/right, front/behind, standing on, close by}$\}$.

\section{SG-Bot: Overview}
\subsection{Problem Definition}
From an initial layout state $\mathcal{S}_0$, the embodied agent is tasked with a sequential transformation of objects towards a desired goal state $\mathcal{S}^*$. 
This transformation is achieved by utilizing sequential motion policies $\mathcal{P}$, guided by sensory observations.

\subsection{Inference workflow} 

\smallskip
\noindent
\textbf{Observation.} 
Given an RGB-D image capturing the initial object layout state $\mathcal{S}_0$, as shown in Fig.~\ref{fig:workflow}.a, SG-Bot first extracts all target objects as nodes $\mathcal{V}(O)$ via an arbitrary object detector, \textit{e.g.}, MaskRCNN~\cite{he2017mask}.

\smallskip
\noindent
\textbf{Imagination.}
The extracted object nodes are constructed as a scene graph $\mathcal{G}$ according to commonsense or user-defined rules, as shown in Fig.~\ref{fig:workflow}.b and explained in Sec.~\ref{sgc}.
Next, we evolve $\mathcal{G}$ to a latent shape-aware scene graph $\mathcal{G}^\beta_z$ with shape priors $\beta$ from the initial scene and learned layout-shape distribution $Z$ mentioned in Sec.~\ref{fine}. Finally, SG-Bot imagines a goal scene $\mathcal{S}^{*}$ conditioned on $\mathcal{G}^\beta_z$ via the shape decoder $\mathit{\Phi}_D$ and layout decoder $\mathcal{L}_D$ of a scene generative model Graph-to-3D~\cite{dhamo2021graph}, where $\mathcal{S}^{*}$ comprises of dense point cloud and corresponding bounding box for each object. 

\smallskip
\noindent
\textbf{Execution.}
Each target object in $\mathcal{S}_0$ is first extracted and represented as the back-projected point cloud from the depth map.
Then, as shown in Fig.~\ref{fig:workflow}.c and explained in Sec.~\ref{matching}, these objects are matched with the corresponding dense point clouds in $\mathcal{S}^{*}$ through iterative registration, \textit{e.g.}, ICP~\cite{besl1992method, zhang1994iterative}.
Based on the outcomes of this registration process, SG-Bot generates per-object manipulation policies $\mathcal{P}_t$ filtered and refined by object occupancy checking at each action step $t$.
SG-Bot continues to iteratively reposition objects in $\mathcal{S}_0$ towards $\mathcal{S}^{*}$ until all objects are effectively rearranged.

\section{SG-Bot: Methodology}
\subsection{Object Extraction}
Given a cluttered scene $\mathcal{S}_0$ as the initial state, SG-Bot first performs semantic instance segmentation to segment all target objects, as shown in Fig.~\ref{fig:workflow}.a.
Specifically, we adopt MaskRCNN to jointly predict the object masks and category labels.
Then, each object is represented as the back-projected point cloud from the depth map.
These objects, denoted as $\mathcal{V}(O) = \{ v_i(o_{i})~|~i = 1,  \ldots, N \}$, are further collected and processed in the following \textit{Imagination} module.
This module aims to generate the desired goal scene by treating these objects as individual scene graph nodes.

After obtaining target objects $\mathcal{V}(O)$, we follow a coarse-to-fine scheme to generate the desired goal scene, which is leveraged to guide the object action.

\subsection{Coarse Stage: Goal Scene Graph Construction}
\label{sgc}
SG-Bot establishes a goal scene graph $\mathcal{G}=\left\{\mathcal{V}(O),\mathcal{E}(\Gamma)\right\}$ via determining the edge type $\gamma_{i\to j} \in \Gamma$ for each edge in $\mathcal{E}(\Gamma)$, as shown in Fig.~\ref{fig:workflow}.b. 
In this paper, two modes are supported to define edges between nodes:

\smallskip
\noindent
\textbf{Commonsense mode}.
Following the recent trend of knowledge representation with graphs~\cite{chen2020review}, we represent common human knowledge in the form of edge attributes $\Gamma$ within a scene graph. 
For instance, for the scene containing a {plate}, the {fork} and {knife} are typically placed to the \texttt{left and right of} the {plate}.
Additionally, the {spoon} needs to be placed  \texttt{in front of} the {plate} if it exists. 
For the case without a {plate}, the {spoon} tends to be placed  \texttt{close by} the {bowl} or {cup}. 
Moreover, other objects need to be placed  \texttt{in front of} the {plate}, {bowl}, and {cup}, etc.
Any unusual objects that appear on the table will be identified as obstacles and subsequently removed, which makes the final $M$ nodes from $N$ elements, $M \leq N$.
Similar rules are naturally introduced based on the category of the object and commonsense. One way to achieve this is to use LLM to choose the optimal relationship according to the provided $\Gamma$.

\smallskip
\noindent
\textbf{User-defined mode}.
In contrast to the uncontrollable \emph{Commonsense mode}, we demonstrate that one of the main advantages of introducing the scene graph representation is that it enables the controllable \emph{User-defined mode}.
Users can manipulate the scene graph from a long-term perspective, e.g., using a GUI, by directly editing the edges and nodes in $\mathcal{G}$ to interact with the edge database $\Gamma$ and nodes.

\subsection{Fine Stage: Graph to Scene Generation}
\label{fine}
SG-Bot stacks the architecture of Graph-to-3D~\cite{dhamo2021graph} to generate a plausible goal scene. Graph-to-3D conditions on the latent shape-aware scene graph denoted as $\mathcal{G}^\beta_z$, which evolves from $\mathcal{G}$ and ensures the coherent shape transformation from the initial scene to the goal scene.

\begin{figure}[t]
    \centering
    \includegraphics[width=\linewidth, angle=0]{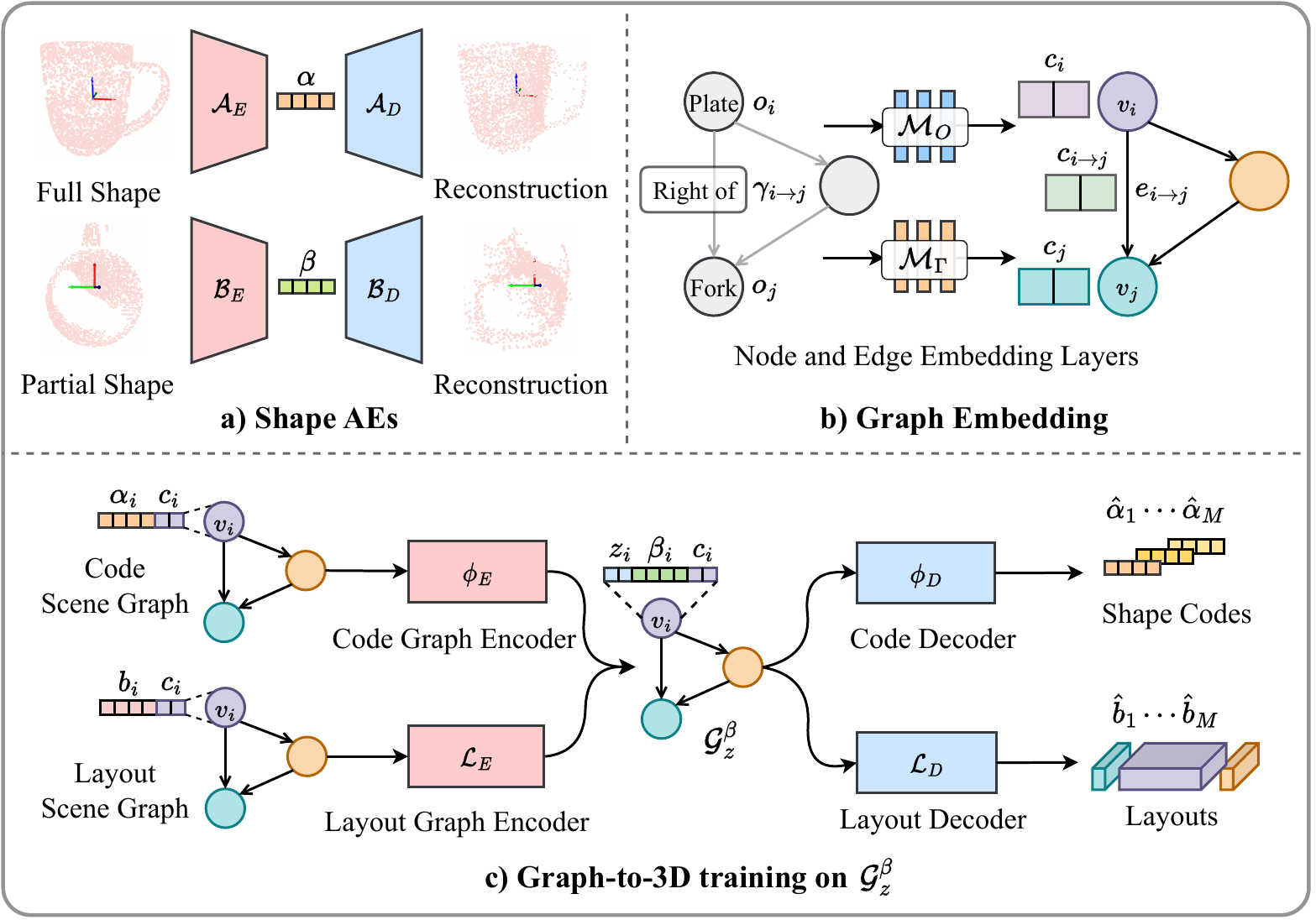}
    \vspace{-15pt}
    \caption{\textbf{Modular Training.} \textbf{a)} $\mathcal{A}_E,\mathcal{A}_D$ are trained using full shapes in the canonical view to have the shape code $\alpha$, while $\mathcal{B}_E,\mathcal{B}_D$ are trained on partial shapes in the initial scenes under the camera view to have the shape priors $\beta$. $\mathcal{A}_D$ and $\mathcal{B}_E$ are retained during inference. \textbf{b)} A scene graph with textual information is processed through embedding layers $\mathcal{M}_O, \mathcal{M}_\Gamma$ to have implicit class features $c_i,c_{i \to j}$ on each node and edge.  \textbf{c)} For training Graph-to-3D on goal scenes, the processed scene graph is first concatenated with $\alpha$ and bounding box parameters $B$ on the shape branch $\mathit{\Phi}(\mathit{\Phi}_E,\mathit{\Phi}_D)$ and layout branch $\mathcal{L}(\mathcal{L}_E,\mathcal{L}_D)$ respectively. $\mathit{\Phi}$ and $\mathcal{L}$ jointly model the layout-shape distribution $Z$~\cite{dhamo2021graph}. This model incorporates $\beta$ from initial scenes to create $\mathcal{G}_z^\beta$, subsequently estimating $\Hat{\alpha}$ and $\Hat{B}$. Modules in \textbf{b)} and \textbf{c)} are jointly trained, with $\mathcal{M}_O, \mathcal{M}_\Gamma$, $\mathit{\Phi}_D$ and $\mathcal{L}_D$ used during inference.}
    \label{fig:train}
    \vspace{-0.5cm}
\end{figure}

\smallskip
\noindent
\textbf{Shape auto-encoders.}
For this purpose, we first train two shape auto-encoder entities $\mathcal{A}, \mathcal{B}$ of AtlasNet~\cite{groueix2018papier} for different usages, as shown in Fig.~\ref{fig:train}.a. We train $\mathcal{A}(\mathcal{A}_E,\mathcal{A}_D)$ with full points under canonical view,  whose encoder $\mathcal{A}_E$ offers shape codes $\alpha$ for training Graph-to-3D after. $\mathcal{B}(\mathcal{B}_E,\mathcal{B}_D)$ is trained with normalized object points under camera view in initial scenes to have initial shape priors $\beta$. The encoder $\mathcal{B}_E$ of $\mathcal{B}$ is preserved to produce $\beta$ during the training of Graph-to-3D and the final SG-Bot workflow. The training process of $\mathcal{A}, \mathcal{B}$ aligns with the original AtlasNet.

\smallskip
\noindent
\textbf{Scene generative model.}
After obtaining $\alpha$ and $\beta$, the training of Graph-to-3D starts with embedding $\mathcal{G}$ shown in Fig.~\ref{fig:train}.b. The category information $c_i \in \mathcal{C}^{node}$ for $i$-th node is obtained by passing its textual information $o_i$ through node embedding layers $\mathcal{M}_O$, while $c_{i \to j} \in \mathcal{C}^{edge}$ is obtained by edge embedding layers $\mathcal{M}_\Gamma$ with $\gamma_{i \to j}$. Based on $\mathcal{G} \mapsto \mathcal{G}=\left\{\mathcal{V}(\mathcal{C}^{node}),\mathcal{E}(\mathcal{C}^{edge})\right\}$, Graph-to-3D, a subsequent dual-branch GCN architecture, is trained by modeling the layout-shape joint distribution $Z$ of goal scenes. As shown in Fig.~\ref{fig:train}.c, in training, the shape branch  $\mathit{\Phi}(\mathit{\Phi}_E,\mathit{\Phi}_D)$ requires the graph to be augmented with ground truth shape codes $\alpha$ in goal scenes as input, whose output $\hat{\alpha}$ is supervised by the same shape codes. In the meantime, the layout branch $\mathcal{L}(\mathcal{L}_E,\mathcal{L}_D)$ takes the scene graph with ground truth bounding boxes $B=\{b_i~|~i=1,.., M\}$ as input and the supervision labels. The two branches interact with each other in the bottleneck to model a latent graph $\mathcal{G}_z$, which shares the same idea of the concept of the latent code in the VAE~\cite{kingma2013auto}. $\mathcal{G}_z = \{\mathcal{V}(z,\mathcal{C}^{node}), \mathcal{E}(\mathcal{C}^{edge})\}$, consisting of $\mathcal{G}$ with sampled $z$ code from the modeled $Z$. More details can be found in~\cite{dhamo2021graph}. Here, we change $\mathcal{G}_z$ as $\mathcal{G}^\beta_z$ by offering each node its shape prior $\beta$ extracted from its counterpart in the initial scene, \textit{i.e.}, $\mathcal{G}^\beta_z = \{\mathcal{V}(z,\beta,\mathcal{C}^{node}), \mathcal{E}(\mathcal{C}^{edge})\}$, to make $\hat{\alpha}$ and $\hat{b}$ aware of initial shapes. 

\smallskip
\noindent
\textbf{Controllable scene imagination.}
After training, we subsequently engage in the process of generating the desired goal scene $\mathcal{S}^*$ conditioned on $\mathcal{G}^\beta_z$, shown in Fig.~\ref{fig:workflow}.b. 
This is accomplished through combination of code decoder $\mathit{\Phi}_D$, shape decoder $\mathcal{A}_D$, and layout decoder $\mathcal{L}_D$:

\begin{small} 
\begin{subequations}
\begin{gather}
S = \mathcal{A}_D(\hat{\alpha}), \quad \hat{\alpha} = \mathit{\Phi_D(\mathcal{G}_z^\beta)}, \quad \hat{\alpha} = \{\hat{\alpha}_i~|~i = 1,...,M\},\\
\Hat{B} = \mathcal{L}_D(\mathcal{G}_z^\beta), \quad \Hat{B}=\{\hat{b}_i~|~i = 1,...,M\},
\end{gather}
\end{subequations}
\end{small}
where $\hat{\alpha}$ denotes the set of estimated shape codes, and $S$ is the set of normalized shapes decoded from $\hat{\alpha}$.
$\hat{B}$ denotes the layout of object bounding boxes in the desired scene $\mathcal{S}^*$. 
$S$ then is transformed and populated into $\hat{B}$ to synthesize $\mathcal{S}^*$.

\subsection{Advantages of Coarse-to-Fine Scheme}
SG-Bot features three key advantages:
First, in the coarse stage, it utilizes a scene graph as an intermediary form of the target scene. This graph allows for multiple relationships between any two objects and enhances natural and intuitive human-computer interaction.
Users can intuitively perceive the spatial distribution of objects within the scene through a 2D graphical scene graph, enabling direct editing through a GUI.
Second, leveraging the scene graph as an intermediate representation allows for the seamless integration of commonsense knowledge, enabling automated scene rearrangement.
Third, in the fine stage, we introduce the generative model to supplement missing fine-grained details, such as object shapes and poses, in the scene graph representation. 
This guides the robot in performing precise operations.

\subsection{Goal-Guided Object Matching and Manipulation}
\label{matching}
After obtaining $\mathcal{S}^*$, SG-Bot performs object matching by point cloud registration and rearranges objects after occupancy check in each round, as shown in Fig.~\ref{fig:workflow}.c, transferring $\mathcal{S}_0$ to $\mathcal{S}^*$. We illustrate the process with the first round:

\smallskip
\noindent
\textbf{Object matching.}
SG-Bot compares $\mathcal{S}^*$ with the initial scene $\mathcal{S}_0$ to calculate the necessary transformation $\mathbf{T}=[\mathbf{R}|\mathbf{t}]$ for each object, where $\mathbf{R} \in \mathbb{R}^{3\times3}$ and $\mathbf{t} \in \mathbb{R}^3$ represent rotation and translation respectively.
Therefore, in this module, the objective can be defined as,

\begin{small} 
\begin{equation}
 \left[\mathbf{R}^*,\mathbf{t}^*\right]=\underset{\mathbf{R},\mathbf{t}}{\text{argmin}}\sum_{i=1}^{N_P}( \underset{q\in Q}{\min}||{\mathbf{R}p_i + \mathbf{t}-q}||^2)+I_{SO(3)}(\mathbf{R}),
 \label{icp}
\end{equation}
\end{small} 
where $\mathbf{R}^*$ and $\mathbf{t}^*$ represent the optimal rotation and translation parameters we aim to find.
$p_i$ denotes one of the $N_P$ points in object $P$ of initial scene $\mathcal{S}_0$.
After transforming $p_i$ from $\mathcal{S}_0$ to the goal scene $\mathcal{S}^*$ with $\mathbf{R},\mathbf{t}$, its corresponding nearest point in $\mathcal{S}^*$ is denoted as $q$ inside object $Q$.
$I_{SO(3)}(\mathbf{R})$ enforces $\mathbf{R}$ should lie in the special orthogonal
group $SO(3)$~\cite{zhang2021fast}.
Since the generated objects in the goal scene are dense and complete, we observe that vanilla ICP can effectively solve the problem in Eq.~\ref{icp} when provided with a well-suited initialization.

Given an object $P$ from the initial scene $\mathcal{S}_0$, its goal location is indicated by the generated object $Q$ in $\mathcal{S}^{*}$.
We initialize the pose $\mathbf{T}$ by first centralizing each point cloud and then uniformly generating candidate rotations.
We represent rotation using angles around the $x$, $y$, and $z$ axes, dividing the interval of each axis's rotation angle [-$\pi$, $\pi$] into $n$ segments, resulting in a total of $n^3$ candidate rotations, where $n=5$ in the implementation. 
Finally, we apply ICP to estimate $\mathbf{R}^*, \mathbf{t}^*$, where $\mathbf{t}$ is initialized with \textbf{0} vector, while $\mathbf{R}$ is initialized with each candidate rotation.
This will result in $n$ outcomes from ICP.
We select the solution that minimizes Eq.~\ref{icp} as the final result.

\begin{figure*}[htp!]
 \centering
 \includegraphics[width=0.8\textwidth, angle=0]{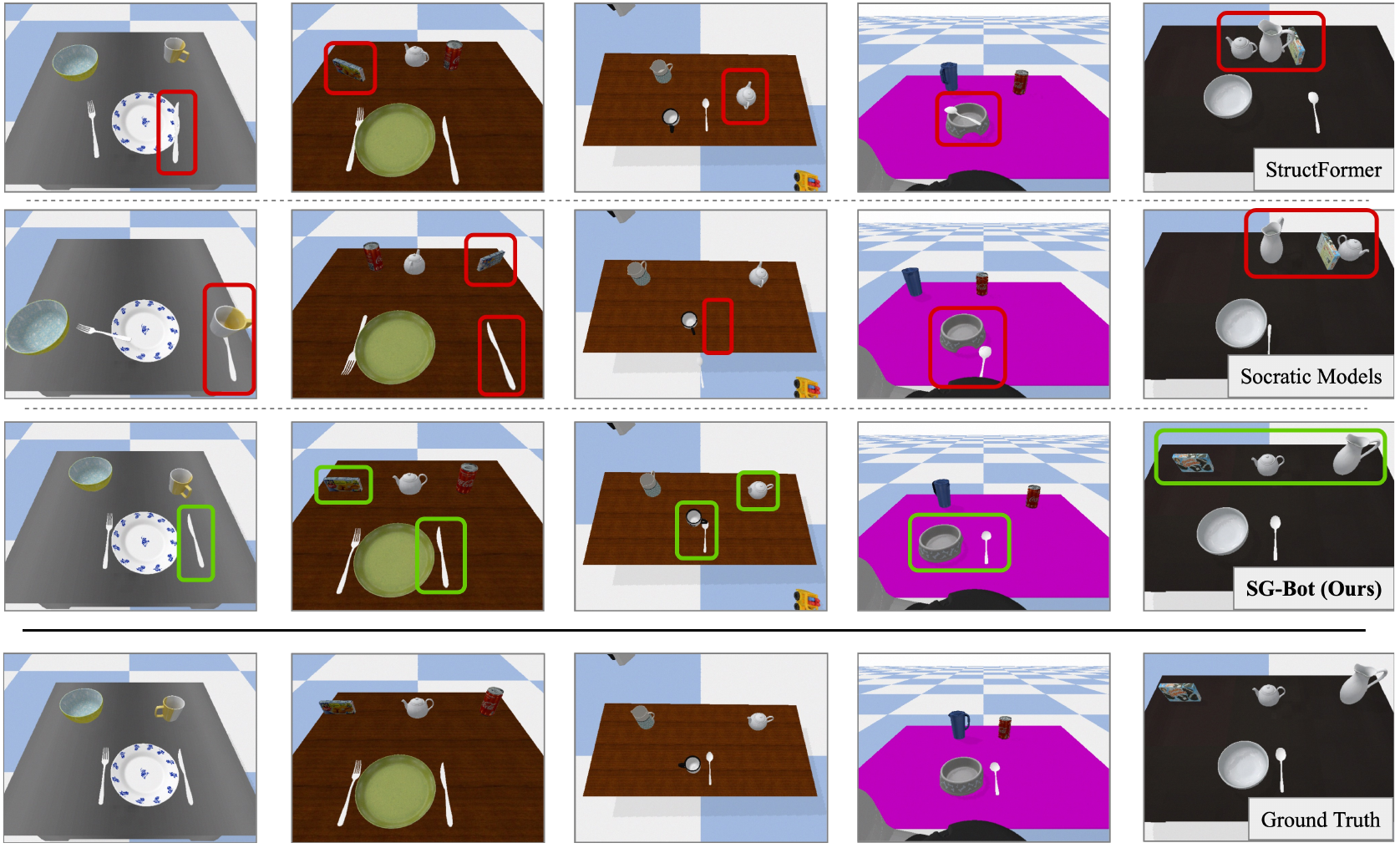}
 \vspace{-5pt}
 \caption{\textbf{Visualization results in simulation.} We compare SG-Bot with state-of-the-art methods StructFormer~\cite{liu2022structformer} and Socratic Models~\cite{zeng2022socratic}. We highlight the superiority of SG-Bot via rectangles.}
 \label{fig:comparison}
 \vspace{-0.3cm}
\end{figure*}%

\begin{table*}[t]
\centering
\caption{Performance evaluation on three aspects -- errors ($rad, cm$), success rate (\%) and fidelity.}
\vspace{-5pt}
\scalebox{1.2}{
    \begin{tabular}{l  c c c c  c c  c c }
     \toprule
        \multirow{2}{*}{Method} & \multicolumn{4}{c}{Rearrangement Errors $(\downarrow)$} & \multicolumn{2}{c}{Success Rate $(\uparrow)$} & \multicolumn{2}{c}{Scene Fidelity $(\downarrow)$}
        \\
        \cmidrule(r){2-5} \cmidrule(r){6-7} \cmidrule(r){8-9}
        & $R_{\text{e}}$ & $t_{\text{e}}$ & $R_{\text{f}}$ & $t_{\text{f}}$ & $\text{IoU}_\text{0.25}$ & $\text{IoU}_\text{0.50}$ & FID & FID-CLIP\\
    \midrule 
        StructFormer~\cite{liu2022structformer} & \textbf{0.28} & 10.58 & 0.18 & 11.17 & 28.03 & 14.01 & 91.46 & 6.32\\
        Socratic Models~\cite{zeng2022socratic} & -- & 12.09 & -- & 13.36 & 43.71 & \textbf{36.58} & 86.46 & 6.96 \\
        \textbf{SG-Bot (Ours)} & 0.38 & \textbf{4.49} & \textbf{0.09} & \textbf{4.61} & \textbf{53.92} & 34.20 & \textbf{58.29} & \textbf{3.91}\\
    \bottomrule
    \end{tabular}
    }
\label{tab:fid}
\vspace{-0.4cm}
\end{table*}

\noindent
\textbf{Object manipulation.}
To determine the final robot action, we select an object $P$ from $\mathcal{S}_0$ and check for occupancy: We measure the point-wise $L2$ distance between its counterpart $Q$ in $\mathcal{S}^*$, and all objects in $\mathcal{S}_0$. If the shortest distance $d$ is smaller than a set threshold $\sigma$, it implies a potential collision. We then bypass moving $P$ and evaluate the next object. This continues until an object with $d > \sigma$ is found, which is then moved to the target pose by its $\mathbf{T}$.

The rearrangement ends in this manner when all objects are in their ideal poses.

\section{Experiment}

\subsection{Implementation Details}

\smallskip
\noindent
\textbf{Dataset.}
We collect a synthetic dataset containing 1,042 realistic initial-goal RGB-D scene pairs with scene graph labels. First, we mix the meshes in Google Scanned Objects~\cite{downs2022google} and HouseCat6D~\cite{jung2022housecat6d} as the object database. Then, we randomly place objects on the tables to render the initial scenes into images using NVISII~\cite{morrical2021nvisii}. The goal scenes are set up using the rules mentioned in Sec.~\ref{sgc}. Then, we construct scene graph labels by comparing the spatial relations of the objects following~\cite{wald2020learning,zhai2023commonscenes}. We define six types of relations as the edge class database $\Gamma$, including spatial, proximity, and support information, representing the \emph{User-defined mode}. 

\smallskip
\noindent
\textbf{Trainval setup.}
We use 952 scenes as the training split and 90 scenes as the validation (test) split. All modules in our pipeline are trained on a single NVIDIA 3090 GPU. We adopt the Adam optimizer with an initial learning rate of 1e-4 to train each module. $\mathcal{A}$ is trained for 500 epochs on the meshes in the training split. $\mathcal{B}$ is trained for 5 epochs in terms of all partial points of each object in the training split. $\mathcal{M}_O, \mathcal{M}_\Gamma, \mathit{\Phi}, \mathcal{L}$ are jointly trained for 600 epochs.

\begin{figure*}[t]
 \centering

 \includegraphics[width=0.95\textwidth, angle=0]{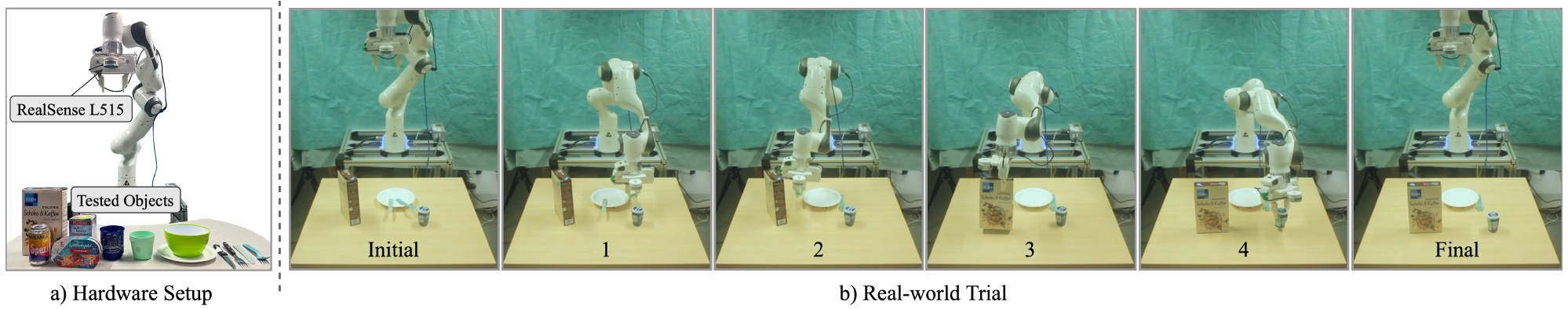}
 \caption{\textbf{Real-world experiment.} a) We tested unseen cross-category objects with a physical manipulator. b) Action decomposition of one trial during the rearrangement.}
 \label{fig:hardware}
 \vspace{-0.4cm}
\end{figure*}%

\subsection{Evaluation Protocols}

\smallskip
\noindent
\textbf{Baselines.}
We reproduce two methods representing different routines on the dataset for the comparison: \emph{First,} StructFormer~\cite{liu2022structformer}, a transformer-based method that autoregressively transforms objects to the goal state based on the current observation and previous states, is fully trained on our dataset. \emph{Second,} Socratic Models~\cite{zeng2022socratic}, a LLM-based method that connects an object detection module~\cite{gu2021open}, GPT~\cite{brown2020language}, and a motion planning method CLIPort~\cite{shridhar2022cliport} in a series, where we use text-davinci-002 for LLM and train CLIPort solely using our dataset. All training and evaluation procedures use the same trainval splits as our method. More details about the reproduction can be found on our project website.

\smallskip
\noindent
\textbf{Metrics.} \emph{First,} for evaluating the rearrangement accuracy, we report the errors of estimated rotation $R_\text{e}$ and translation $t_\text{e}$ comparing final positions with ground truth following~\cite{liu2022structformer}. We also report the errors of final poses $(R_\text{f}, t_\text{f})$, as the final states of rearrangement are slightly different from the predicted ones because of the table-object physical interaction. \emph{Second,} for the rearrangement success rate, we calculate the IoU between the bounding boxes of rearranged and ground truth objects. If IoU $>\sigma$, it counts as a success, $\sigma = 0.25, 0.50$. Note that this is a strict metric, as objects tend to be tiny, where even a small misalignment can cause failure. \emph{Additionally,} inspired by some research on indoor scene synthesis~\cite{ritchie2019fast,paschalidou2021atiss,zhai2023commonscenes}, we believe that measuring the fidelity of the rearranged scene is critical for evaluating global performance. For this, we render rearranged scenes of all methods and ground truth scenes under a specific viewpoint, and then we employ the commonly adopted Fréchet Inception Distance (FID)~\cite{heusel2017gans} and recent FID-CLIP~\cite{kynkaanniemi2022role}.
\begin{figure}[t]
 \centering
 \includegraphics[width=1\linewidth, angle=0]{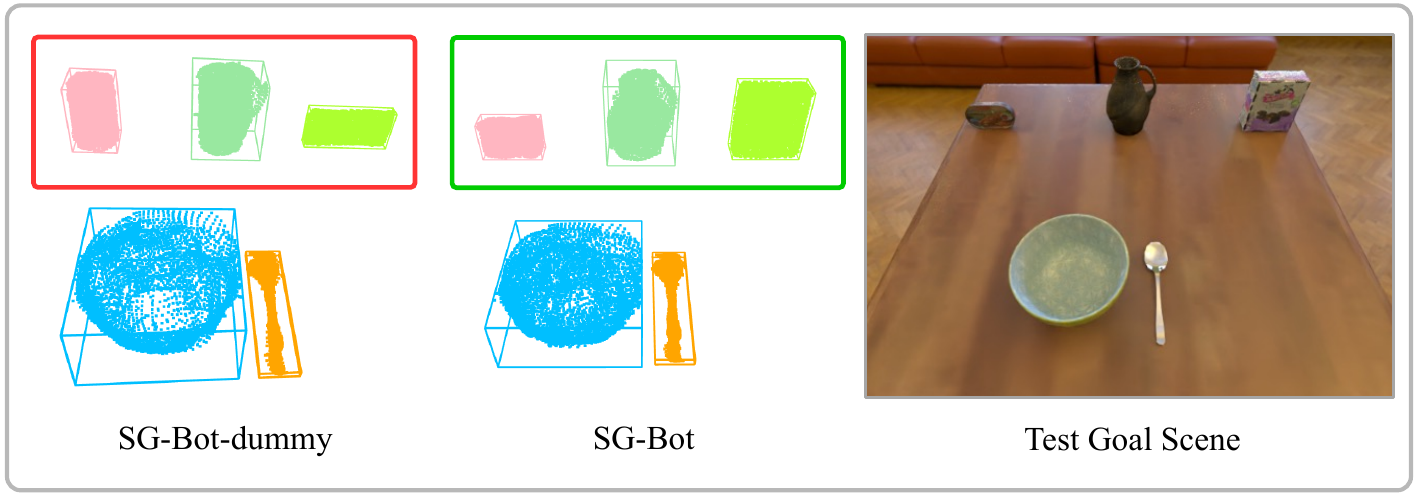}
 \vspace{-15pt}
 \caption{\textbf{Functional shape priors.} Without shape priors, SG-Bot-dummy generates inconsistent shapes \textbf{(left)}. SG-Bot controls the generated shapes close to the ground truth (right) with the help of initial shape priors \textbf{(middle)}.}
 \label{fig:ablation}
 \vspace{-0.4cm}
\end{figure}%
\subsection{Simulation Experiments}
We import meshes with their initial poses to a PyBullet environment~\cite{coumans2021} to evaluate each method. In the simulation, we leverage ground truth instance masks and remove the effect of the robotic low-level control.

\smallskip
\noindent
\textbf{Quantitative results.}
As shown in Table~\ref{tab:fid}, our method surpasses the previous approaches on most metrics by a large margin. SG-Bot obtains lower rearrangement errors on the final states and yields competitive success rates, indicating that SG-Bot shows more accurate object-level rearrangement. For instance, SG-Bot decreases 50.0\% on $R_\text{f}$ and 58.7\% on $t_\text{f}$ compared with StructFormer~\cite{liu2022structformer}. When using $\text{IoU}_{0.25}$, SG-Bot increases 10.21\% on success rate compared with Socratic Models~\cite{zeng2022socratic}. On the scene-level comparison, SG-Bot shows more fidelity in rearranged scenes than other methods, modeling a more similar scene distribution to ground truth supported by lower FID and FID-CLIP.

\smallskip
\noindent
\textbf{Qualitative results.}
We show several qualitative comparisons of rearranged scenes in Fig.~\ref{fig:comparison}, where our method shows clear advantages against others. For example, in the first scene, the rearranged knife collides with the plate or the cup in StructFormer and Socratic Models, which is better placed with our method. In the last scene, our method can separate objects at a sensible distance while others make them unevenly distributed. 

\smallskip
\noindent
\textbf{Ablation study.} We ablate the shape priors, resulting in \emph{SG-Bot-dummy}, a framework only taking the original latent scene graph $\mathcal{G}_z$. As shown in Fig.~\ref{fig:ablation}, SG-Bot powered by $\mathcal{G}_z^\beta$ has more controllable ability than SG-Bot-dummy, generating more consistent shapes to the objects in the scenes. We also report quantitative comparisons in Table~\ref{tab:ablation}.

\begin{table}[t]
\centering
\caption{Ablation -- errors ($rad, cm$), success rate (\%) and fidelity.}
\vspace{-5pt}
\scalebox{0.9}{
    \begin{tabular}{l  cc cc cc }
     \toprule
        \multirow{2}{*}{Method} & \multicolumn{2}{c}{ Errors $(\downarrow)$} & \multicolumn{2}{c}{ Success Rate $(\uparrow)$} & \multicolumn{2}{c}{Scene Fidelity $(\downarrow)$}
        \\
        \cmidrule(r){2-3} \cmidrule(r){4-5} \cmidrule(r){6-7}
        & $R_{\text{f}}$ & $t_{\text{f}}$ & $\text{IoU}_\text{0.25}$ & $\text{IoU}_\text{0.50}$ & FID & FID-CLIP\\
    \midrule 
        SG-Bot-dummy & 0.09 & 4.86 & 46.32 & 27.08 & 64.28 & 4.20 \\
        SG-Bot & 0.09 & \textbf{4.61}  & \textbf{53.92} & \textbf{34.20} & \textbf{58.29} & \textbf{3.91} \\
    \bottomrule
    \end{tabular}
    }
\label{tab:ablation}
\vspace{-0.5cm}
\end{table}

\subsection{Real-world Experiments}
We test SG-Bot in real-world scenarios using a 7-DoF Franka Panda robot with a parallel-jaw gripper as the end-effector. The sensor mounted on the gripper base is a RealSense L515 RGB-D camera. The framework is run on an NVIDIA 3080 laptop GPU. Different from the strategy in the simulation, we use Contact-GraspNet~\cite{sundermeyer2021contact} to generate appropriate grasps on each masked object and rearrange them by reasoning the relative pose and executing the best grasp with Moveit!~\cite{coleman2014reducing}. We show an example work stream in Fig.~\ref{fig:hardware} out of 5 rounds where we test with unseen objects. More trials can be found on the project website. Our method can still maintain the rearrangement performance consistent with the one in the simulation.
\section{CONCLUSIONS}
In this paper, we present a novel robotic rearrangement framework, \emph{SG-Bot}, which follows a three-phase procedure: observation, imagination, and execution to handle this task.
With its unique coarse-to-fine design, SG-Bot embraces the synergy of commonsense priors and dynamic generation capabilities, all within a lightweight, real-time, and customizable pipeline.
Extensive experiments on both simulation and real-world datasets demonstrate the superiority of SG-Bot.
Future work will explore deformable point cloud matching for enhanced accuracy or accelerated point cloud
alignment~\cite{malis2023complete}.

\section*{Acknowledgement}
\textbf{We are truly grateful for the reviews provided!} Due to the page limit, we are not able to add more content, but we are very open to and feel excited for further discussions! We would like to thank Mr. Shun-Cheng Wu for the early discussion. We also would like to thank Ms. Chang Gao (open to graphic design jobs, email: \texttt{gaochang960605@gmail.com}) for the teaser design.


\bibliographystyle{IEEEtran}
\bibliography{IEEEabrv,literature}
\end{document}